\documentclass[12pt,oneside]{article}
\usepackage{authblk}
\usepackage{graphicx}
\usepackage{cite}
\usepackage{url}
\usepackage[colorlinks=true,linkcolor=blue,citecolor=blue,urlcolor=blue]{hyperref}
\usepackage{geometry}
\title{\textbf{ The wall confronting large language models}}
\author[1,2,3]{P.V. Coveney}
\author[4,5,6]{S. Succi}

\affil[1]{\footnotesize Centre for Computational Science, University College London, London WC1H 0AJ, U.K.}
\affil[2]{\footnotesize Advanced Research Computing Centre, University College London, London WC1H 0AJ, U.K.}
\affil[3]{\footnotesize Institute for Informatics, Faculty of Science, University of Amsterdam, 1098XH Amsterdam, The Netherlands}
\affil[4]{\footnotesize Italian Institute of Technology, Viale Regina Elena 291, 00161 Rome, Italy}
\affil[5]{\footnotesize Physics Department, Harvard University, Cambridge USA}
\affil[6]{\footnotesize Mechanical Engineering Department, University College London, London WC1H 0AJ, U.K.}

\begin{document}
\maketitle

\begin{quote}
{\footnotesize\it
\section*{Abstract}

We show that the scaling laws which determine the performance of large language models 
(LLMs) severely limit their ability to improve the uncertainty of their predictions.  
As a result, raising their reliability to meet the standards of scientific inquiry is intractable
by any reasonable measure.
We argue that the very mechanism which fuels much of the learning power  
of LLMs, namely the ability to generate non-Gaussian output distributions 
from Gaussian input ones, might well be at the roots of their propensity to produce error pileup, ensuing information catastrophes and degenerative AI behaviour.
This tension between learning and accuracy is a likely candidate mechanism
underlying the observed low values of the scaling components.
%\cite{Hammond} 
It is substantially compounded by the deluge of spurious correlations pointed 
out by Calude and Longo which rapidly increase in any data set merely as a function 
of its size, regardless of its nature.
The fact that a degenerative AI pathway is a very probable feature of 
the LLM landscape does not 
mean that it must inevitably arise in all future AI research. 
Its avoidance, which we also discuss in this paper, necessitates putting a much higher 
premium on insight and understanding 
of the structural characteristics of the problems being investigated.  
}
\end{quote}

\newpage 

\section{Introduction}

In the last decade AI and most notably machine learning (ML)
have taken science and society by storm, with relentless progress 
reported, most recently in the context of generative AI's large language models (LLMs) 
which display impressive capabilities in natural language processing that are
not infrequently claimed to exceed their human counterpart. 
Such developments purport to adumbrate a completely unprecedented approach to 
a class of scientific problems whose complexity lies
beyond reach of our best theoretical and computational
models based on the scientific method, models that offer deep
insight into and understanding of the basic mechanisms underlying the phenomena
under investigation. 
AI's success stories are the talk of the decade: chess and GO victories, 
self-driving cars and
AlphaFold's prediction of protein structures stand out 
as some the most spectacular cases in point \cite{Jumper},
even dominating the Stockholm podium both in chemistry and physics in 2024 \cite{Nobel2024}. 
However, at root, ML is based on a mathematical black-box procedure which 
is entirely unaware of the underlying physics, although adding physics-informed constraints 
often improves its convergence behaviour \cite{Raissi}.  

The scientific and wider societal implications are paramount. 
Here we pinpoint precisely this aspect, with specific focus on the sustainability of 
LLMs applications based on a simple assessment of their scaling laws. By now it is common knowledge that the small number of AI tech companies capable of competing in the LLM domain have an insatiable desire for electrical energy and the electrical power that comes with it, to the point of re-opening and planning to build many nuclear reactors adjacent to their data centres where exceptionally large GPU-accelerated supercomputers reside. These resources are being used to build so-called “frontier models”, LLMs comprised of trillions of parameters (the connection weights in the underlying neural networks). However, although these companies guard their AI capabilities behind secure firewalls and do not share any of the technical details of their commercial products, it is clear that the improvements accruing are relatively limited. So far as the diminishing returns of large LLMs are concerned, a good example is GPT-4.5. Its parameter count is speculated to be in the range of 5-10 trillion \cite{interconnects2024, wandb2024}, and it is probably based on a mixture-of-experts architecture with {\it ca} 600 billion active parameters during inference. The API cost is 15 to 30 times higher than that of GPT-4o, its predecessor, which is indicative of the increase in model size.

The cost comparison is even more extreme when measured against smaller distilled models (see \cite{openaiPricing}). In contrast to “reasoning”-tuned models like o3, GPT-4.5 relies on sheer scale and pre-training rather than chain-of-thought fine-tuning with reinforcement learning. This yields qualitative gains in subjective areas (such as its abilities in writing and expressing empathy) but nothing substantial in verifiable domains such as mathematics and science \cite{bharath2024,reddit2025}. Interestingly, OpenAI initially described GPT-4.5 as “not a frontier model”, but later retracted that statement. The public only has access to the preview version.  Another example is Meta’s Llama 4 Behemoth, a 2-trillion parameter model that appears to underperform relative to its scale, based on community sentiment and the release delay \cite{interconnects2024, wandb2024}. 

Albeit elementary in many respects, the consequences of the very low scaling exponents
do not appear to have been explicitly spelled out in the scientific literature, which 
is precisely the gap we address in this paper. In short, they underpin the poor peformance of LLMs and the very marginal improvement in their capabilities when trained on larger data sets.
Moreover, we also propose a theoretical scenario supporting the emergence
of such low scaling exponents and relating it to recent findings on the information collapse
of self-trained AI models and the recognized lack of quantitative accuracy of modern LLMs.
Mitigating scenarios are also outlined, underlining the importance of insight
and understanding over purely empirical (\textit {ad hoc}) control.   

\section{Scaling laws in computer simulation}

Computer simulation is an essential component of scientific 
investigation and it is no exaggeration to state that no field of science 
and engineering would look the same without it.
Among its strengths is the fact that all the data and parameters which enter such 
models bear a direct scientific meaning based on our understanding of a 
given domain, often straddling across different disciplines \cite{HEMO, debezenac2017deep, kashinath2021physics, lagergren2020biologically, sel2023physics}. 
This is very different from AI where vast numbers of parameters are in play from the outset in 
artificial neural networks, whose role is solely to fit input to output data through an extensive 
and costly training period. Insight and understanding take a back seat as do the resources 
required to solve any particular problem. 
By contrast, scientific simulation methods are always developed hand in hand with well defined 
understanding about both their computational complexity and the way their errors scale 
with the computational resources required to solve the problem at hand.

As a first concrete example, let us consider the grid discretization 
of a differential equation in $d=1$ spatial dimensions. 
For a discretization method of order $p$
accuracy, the error of the simulation scales as follows
\begin{equation}
\label{EN}
\epsilon_N = c_p N^{-p} \equiv (N/N_c)^{-p}
\end{equation}
where $N$ is the number of mesh nodes in the grid and $c_p$
is a numerical prefactor dictating the critical size 
$N_c=c_p^{1/p}$ above which the error decreases below one.
Scaling is usually associated with exponents since they dictate the trend
of the error in the ``thermodynamic limit" $N \to \infty$ as the size of the problem grows.
However, real world applications--even the largest ones--take place in 
finite-size spaces, hence the prefactors play a significant role too, as 
they control the ``effective thermodynamic limit", $N \gg N_c$.
A simple example clarifies the point.
Consider the error associated with a centered finite differencing of
the first order spatial derivative
$$
\frac{df}{dx} \sim \frac{f(x+h)-f(x-h)}{2h} 
= f'(x) + f^{'''}(x) \frac{h^2}{6} + O(h^3).
$$
The relative error is $|f^{'''}(x)/f'(x)| h^2/6 \sim (h/l)^2$
where $l$ indicates the smallest lengthscale of the function 
$f(x)$. Let $h=L/N$, $L$ being the size of the domain; after some algebra, we finally obtain for this error the expression
\begin{equation}
\epsilon_N \sim (L/l)^2N^{-2}, 
\end{equation} 
indicating second order accuracy with $N_c = L/l$.
This clearly shows that localized functions, with $l/L \ll 1$ and
extended functions with $l/L \sim 1$, both fall under second order
scaling, but the latter show much smaller errors than the former. 
Hence, in actual practice, good convergence implies not only 
large negative exponents but also small prefactors.

For the $d>1$ dimensional case, with a number $G=N^d$
of mesh points, the very same error reads as follows:

\begin{equation}
\label{EG1}
\epsilon_N = c_p G^{-p/d}.
\end{equation}

This indicates a rapid deterioration of the accuracy with increasing 
dimension, one face of the notorious ``curse of dimensionality" (CoD) according to 
which the cost of computing properties of interest scales exponentially in the 
dimensionality of the problem of interest.   

It is thus of major interest to analyse the scaling of the error with the cost
of the simulation, which we take here to be the number of operations required to
complete the task. For a local grid method the cost scales as
$C \sim N^{d_4} = G^{d_4/d}$, where $d_4$ is the computational
dimension of 4-dimensional spacetime, namely $d+1$ for propagation phenomena
and $d+2$ for diffusion.
Hence we obtain:
\begin{equation}
\label{EC}
\epsilon_C = c_C G^{-p/d_4}
\end{equation}
which leads to an even smaller exponent.
For systems involving all-to-all interactions between points on the grid, such 
as gravitation, the scaling is significantly worse, with $C \sim G^{2+1/d}$ leading to $d_4=2d+1$.

The curse of dimensionality is the main reason why grid methods are generally used only
up to around three dimensions, which is fortunately that of the physical world we live in.
Yet, this is very short of addressing the requirements of many problems
in modern science and engineering, especially but not exclusively
those in the quantum realm, which must often deal with ultra-high-dimensional
spaces with thousands of dimensions or more \cite{Poggio}.
This is the reason why high-dimensional problems in $d>3$ are usually
handled by stochastic techniques, such as the Monte Carlo (MC) method.
By suitably sampling the relevant regions of state space, the 
convergence of the MC method scales as the inverse square root of the 
sample size, that is:

\begin{equation}
\label{EG2}
\epsilon_{MC} = c_{MC} G^{-1/2}
\end{equation}

Comparing with eq.(\ref{EG1}), it is readily seen that
MC outperforms grid methods under the condition 
$1/2 \ge p/d$, that is 

\begin{equation}
\label{D2P}
d > 2p
\end{equation}
For the typical case $p=2$, this delivers $d>4$, which indeed matches the common practice in computer simulation.

Hyperdimensional systems offer possibly the most advantageous terrain for AI 
by comparison with conventional computer simulations, precisely because AI techniques seem able 
to search out sparse solutions in very high dimensional spaces \cite{Poggio}.
However, our principle point, as the ensuing detailed inspection 
reveals, is that this terrain of operation comes at a very high computational price.

\section{Scaling of Large Language Models}

The convergence of modern ML schemes in general, and LLMs in
particular, is much harder to assess than for computer simulation
methods, since they are based on a combination of deterministic and stochastic 
techniques in ultra-high-dimensional spaces with $d \gg 3$. 

A milestone result in LLM research was the discovery that LLMs
not only survive overfitting (more parameters than data) but also
seem to display ``emergent" properties, i.e. they learn better as they
get bigger. These scaling properties bear paramount technological 
implications, as they provide the basic booster for the relentless current
big-tech race towards larger and larger LLMs, with parameters now extending well into
the range of trillions.
On a loose but telling note, this is still three decades short of the number of 
neural connections in the human brain, $10^{15}$, and yet they consume some
one hundred million times more power (GWatts as compared to the very modest 20 Watts required by our brains).

Given that such gargantuan numbers demand nuclear power plants 
to meet the needs of the modern AI industry as of now, it is well worth
assessing the scaling performance of LLMs in more detail. 
The question is plain: what return are we getting for such an insatiable demand for energy? 
%{\bf Note: the number of neurons in 
%($10^{11}$); the number of connections is $10^{15}$, so we are not quite 
%comparing eggs with eggs here} 

In their 2020 paper, the OpenAI team published a very detailed analysis of the scaling laws of 
LLMs in terms of the dependence of the error (assumed here to be equal to the ``loss function", but see our qualifying comments below) 
and processing power on the quantity of data and parameters \cite{Hoffmann}. For the purpose of comparison, let us cast the LLM results as follows

\begin{equation}
\label{D2P1}
\epsilon_K = \left( \frac{N}{N_{cK}} \right)^{-\alpha_K}
\end{equation}
where we identify the loss function with the discretization
error and the number of parameters with the number of 
unknowns in the computer simulation. $K=P,D,C$ stands for ``parameters", ``data" and ``cost", respectively.
As previously mentioned, we take a reference value for these exponents of $\alpha = 0.1$ and $N_c = 10^{14}$ for the critical scale. The corresponding exponents \cite{Chibbaro2017} lie systematically below $0.1$, which 
we take as a (generous) estimate to set up a comparison 
with the convergence exponents and prefactors described in the previous section.
More precisely, the values of these quantities are $\alpha_P=0.075,N_{cP}=8.8 \times 10^{13}$
for the scaling of the loss function with the parameters, 
$\alpha_D=0.095,N_{cD}=5.4 \times 10^{13}$ for the dependence on the 
data and $\alpha_C = 0.05, N_{cC}= 2.3 \times 10^8$ for the computational cost.  

The apparently good news brought about by LLM scaling is basically twofold: first, 
there is no evidence of broken scaling due to overfitting over nearly
six decades; second, the exponent $\alpha \sim 0.1$ is far above 
the value $p/d$ of any grid method in ultra-high-dimensional space: taking $p=2$, any 
$d>20$ would give a lower scaling exponent.
Likewise $G_c \sim 10^{14}$ in $d=20$ dimensions implies 
$N_c = L/l \sim 10^{14/20} \sim 5$, which points to a comparatively
smooth target function.
In other words, a text of 100 words, each 10 letters long, is 
represented by a vector in a $d=1000$ dimensional space. 
In this case, even assuming a grid method could access sufficient memory, which is 
clearly not the case, in order to feature a scaling exponent of 
$0.1$ a grid method would need to work with $p=100$ order accuracy, something 
completely unimaginable.

By contrast, a Monte Carlo simulation of one hundred rigid molecules, each featuring
12 degrees of freedom, tracing a trajectory in a $d=1200$ dimensional space, fits
well within the capabilities of current supercomputers (caveat: MC simulations
usually target average properties and are not supposed to capture single trajectories).
The reason is that MC simulations successfully combat the curse of dimensionality by focussing 
on the degrees of freedom where they are most needed, avoiding wasting
resources on uninteresting regions of state space.     
This requires the use of importance sampling techniques to identify
the hot-spots within the state space. Just to give the reader an idea of how tiny 
these hot spots are, it suffices to remind oneself that a random move in the 
state space of 100 spherical molecules has one 
chance in $10^{260}$ to hit the target \cite{FS}.  
Returning to LLMs, the overwhelmingly bad news is that in 
absolute terms an exponent of $0.1$ is far too low to yield computationally 
tractable improvements based on scaling up the size of the systems now being trained.

\subsection{Transformers as complex dynamical systems}

This is not surprising, given that LLMs employ stochastic
techniques with presumably strong non-Gaussian fluctuations, hence
exposed to a much stronger Resilience of Uncertainty (RoU) 
than in the Gaussian world (for a precise definition of RoU, see Appendix I).
The fact that LLMs use stochastic techniques to predict the next word out of
a pre-existing sequence is well known, but that does not prove that such stochastic
processes are non-Gaussian. 
Very recent work shows that random rectified linear unit (ReLU) 
networks are indeed non-Gaussian processes \cite{Parhi}).   
One can, however, adduce both specific and general arguments to establish that this must indeed be the case. 
The specific one is that LLM architectures are centered upon transformers, whose
basic action is to turn an input signal $x$ into an output $y$.
The distinctive feature of transformers is that the number of neurons is the same
across layers, which strongly facilitates the analogy with discrete dynamical systems.
To highlight this analogy, it proves convenient to cast the transformer update 
as follows:

\begin{eqnarray}
\label{eq:dzdt}
\frac{dz}{dt}  &=& -\alpha [z-f(Wz-b)] \\
\label{eq:dWdt}
\frac{dW}{d\tau}  &=& -\frac{\partial \mathcal{L}}{\partial W}
\end{eqnarray}
with boundary conditions $z(t=0)=x$ and $z(t=T)=y$.
In the above  $x$ (input), $b$ (bias) and $y$ (Output) are $N$-dimensional arrays, 
$W$ is a $N \times N$ weight matrix, $f$ is the activation function, $\mathcal{L}$
is the loss function and $\alpha$ is a relaxation parameters \cite{SS25}.
Note that $t$ represents the ``fast time" across the layers, whereas $\tau$ denotes
the ``slow time" associated with the backward error propagation step. 

A word of caution is in order. In the aforementioned discussion, we have assumed that the Loss function is equivalent to an approximation error. This association is not entirely correct because
an approximation error equal to zero means perfect convergence to the
desired target, whereas LLM practice informs us that pushing the loss function
below a given threshold may be detrimental to the quality of the results (due to overtraining).

In this sense, the loss function is not a metric of the quality of an LLM’s predictions in the same way that an
approximation error is in a ``conventional" numerical scheme; instead, it is a pseudo-metric,
which evades the convergence criteria for errors commonly adopted in scientific computing.
There is nothing wrong with the loss {\em per se}, but given that it is not obvious that zero loss is the most desirable target, a decreasing trend in the loss function with increasing
computational resources cannot necessarily be taken as an indication of success in its predictions. LLM loss is just a self-supervised learning signal. Slight mispredictions of the next word are often inconsequential, especially if they occur in intermediate reasoning traces that can be revised. In fact, driving the loss too low can lead to overfitting or mode collapse.

As such, loss is a rather slippery and ill-defined concept whose lack of equivalence with convergence of errors would melt away were LLMs forced to perform quantitative scientific and mathematical tasks rather than being evaluated based on their qualitative alignment with meaning in a language context.  A very serious scientific drawback of AI and LLMs in particular is the pace of their development, with the claims of awesome capability being superseded by new paradigms which are asserted to be far better before anyone has been able to make sense of how the original algorithm behaves. At the time of writing, large reasoning models (LRMs) are in the ascendent, said to be better still because they provide evidence of how they reach their answers — to say nothing of agentic AI which is supposed to perform better still by orchestrating the interactions between multiple LRMs. This article only considers LLMs. Indeed, through their design rules, LRMs actually render the loss function an even more tenuous measure of their performance. LLMs themselves are all of a kind: they are all based on the original transformer architecture dating back some eight years, and they all perform in similar ways, revealing a remarkably limited underlying diversity in design. Examples of their shared limitations include the fact that they are not very good at making jokes and have an extraordinarily limited repertoire of only two, while when asked to name a random number between 1 and 50, Anthropic’s, Google’s and OpenAI’s LLMs all come up with 27. xAI’s Grok produces 42 first, and then 27 \cite{xia2025private}. In short, they lack {\em novelty}.

For instance, it is not implausible to speculate that the ill-defined nature of the loss function as a metric of success
plays a non-negligible role in the ubiquity of so-called ``Potemkins”\footnote{For the record, the term ``Potemkins" stems from the Russian 
Grigory Potemkin, a field marshal and former lover of Empress Catherine II, who constructed phoney portable
settlements solely to impress the Empress during her journey to Crimea in 1787. The structures would be disassembled after she passed, and re-assembled farther along her route where she could see them again.}, namely 
"the illusion of understanding driven by answers irreconcilable with how any human would interpret a concept”, quoting from \cite{mancoridis2025potemkin}. 
The authors of \cite{mancoridis2025potemkin} go on to state: ``We also find that these failures reflect not just incorrect understanding, but deeper internal incoherence in
concept representations”.  One can associate Potemkins with a general sort of ``holographic accuracy", meaning by this
a convergence which appears to be satisfactory only as long as one does not start scratching beneath the surface. Holographic accuracy 
is possibly a general trait of modern society over and beyond LLMs.

Returning to the transformation in eqns~(\ref{eq:dzdt}) and (\ref{eq:dWdt}), this is then iterated across the entire set of layers 
of the feed-forward network, the analogue of the trajectory of the associated 
dynamical system (once time is made discrete, see \cite{SS25}).

For the sake of concreteness, an explicit example with $N=2$ is provided 
in Appendix II.
For the general case, suppose that input data are sampled from a multivariate
distribution $p_X(x)$; the output distribution is then given by 
\begin{eqnarray}
\label{PXY}
p_Y(y) = \frac{p_X(x)}{|J|(x)} 
\end{eqnarray}
where $|J|$ is the determinant of the Jacobian of the transformation $y=f^L(Wx-b)$ and
$f^L$ indicates repeated application across $L$ hidden layers.
Using the fact that $J=\prod_{i=1}^N \lambda_i$, $\lambda$ being the 
eigenvalues of $J$, the above relation can also be written as:
\begin{equation}
\label{PXYL}
p_Y(y) = p_X(x) \;e^{-\Lambda(x)} 
\end{equation}
where we have set 
$\Lambda(x)= \sum_{i=1}^N log|\lambda_i|(x)$.
The above expression highlights the crucial role of the eigenvalues of 
the Jacobian in shaping the output distribution.
It is therefore clear that even if the input distribution
is Gaussian, the output distribution  will not be so in general (see Appendix II). 
Obviously, the argument is further reinforced
by iteration across multiple layers 
(which is where much of the strength of deep learning comes from).  

A transformer turning Gaussian inputs into Gaussian outputs would feature
very limited learning capability, hence it is natural to associate learning
with the ability to generate non-Gaussian outputs from Gaussian inputs. 
In the following, we shall argue that this very learning capability is
inherently exposed to potentially catastrophic failures and DAI behaviour.
In the above the word ``potential" means that such catastrophic paths
are not inevitable and in the following we shall also provide 
a few simple heuristics to facilitate their avoidance. 
   
To this end, the first observation is that in complex systems, and LLMs certainly qualify as such,
non-Gaussian fluctuations are the rule rather than the exception. 
This is hardly a coincidence: non-Gaussian fluctuations feed the long-range spacetime correlations
which lie at the very heart of complexity (telling it apart from pure chaos and randomness).
This is a well-known hallmark of extended nonlinear systems far from equilibrium
and plays a vital role in most dynamic structure formation processes \cite{Balescu}.
It is therefore entirely reasonable to propose that such long-range correlations
play a vital role in shaping the very structure of LLMs (which, we recall are discrete
dynamical systems), including their alleged ``emergent properties" \cite{SS19,SS22,SS25,CARSON}.

%Incidentally, one may also note in passing Zipf's law
%in linguistics, according to which the frequency of words scales with their inverse
%rank of appearance, a power-law which also points in the same direction. 

The famous ``attention" mechanism, a landmark of LLM research \cite{Vaswani}, 
is itself an outstanding example in point. (As an aside, it is interesting to note
that, when prompted with the question as to a physical analogue of ``attention", ChatGPT 
itself suggests a nonlocal memory mechanism.) 

A second well-known fact about complex dynamical systems is that non-Gaussian fluctuations
often take the form of fat-tailed distributions, such as stretched exponentials and
power-law distributions. The implications for accuracy cannot be understimated, since it
is equally well-known that at a given level of accuracy, sampling fat-tailed distributions requires 
far more data as compared to their Gaussian counterparts, a phenomenon that we name
{\it Resilience of Uncertainty} (RoU). In fact, information catastrophes have been
specifically attributed to the inability to reproduce the tails of the distribution
under self-training scenarios \cite{Shuma}.
Based on the above, it is quite natural to propose that the 
very mechanism fuelling LLMs' ability to learn also exposes them 
to error pileup and a resulting information catastrophe, to be discussed 
further below. 

At this stage a few quantitative comments on the expression (\ref{PXY}) are in order.
First, if the activation function is linear, the Jacobian is a constant and
the output distribution is just a rescaled version of the input distribution: 
Gaussian inputs yield Gaussian outputs, the aforementioned low-learning scenario.
This highlights the key role of nonlinearity in learning in general.
The second observation is that the specific nature of the activation 
functions is also key. Take for instance the hyperbolic tangent $f(z)=tanh(z)$:
at low input $|z| \equiv |Wx-b| \ll 1$, it returns $f(z) \sim z$, hence linear behaviour.
This means that ``small" signals are delta-like distributed around $z=0$ namely 
around $x = W^{-1} b$ (assuming an invertible weight matrix).  
``Large" signals, $|Wx-b|>>1$, on the other hand, saturate the activation function
hence they are peaked around $z \pm 1$, yielding a bimodal distribution (see Appendix II).
These simple heuristics highlight the role of the weights as a sort
of ``effective temperature" of the distribution: small weights lead to
a narrow Gaussian centered about $z=0$, while large weights give a bimodal distribution,
suggestive of a phase-transition. 
Intermediate values provide flat distributions very far from Gaussian behaviour, hence
potentially being exposed to error pile-up on the aforementioned ``large signals".
In passing, we note the beneficial effects of ``small" weights in taming long tails.
However, practical transformers do not generally use $tanh$ activation functions but
ReLU and smooth variants thereof, such as GeLU (Gaussian Error Linear Units).
This family of activations functions leaves positive signals 
intact (linear behaviour) and suppresses negative ones, thus leading to 
quasi ``half-Gaussian" distributions.
This appears to be wise as it protects the large signals from the onset of long tails. 

Summarizing, despite their simplicity, the above considerations illustrate the
delicate tension between the tendency for non-Gaussian behaviour to fuel learning and
the need to inhibit error pileup.
The outcome of such a delicate balance is highly detail-dependent, imposing stringent
constraints on the quantitative predictive power of a general theoretical analysis of this kind.

Overall, this scenario provides a very reasonable explanation for the very
small scaling exponents as a direct consequence of the aforementioned trade-off
between learning and accuracy.   It also provides guidelines for how to attempt to avoid degenerative AI behaviour. However, an important but astonishingly little known property of large data sets needs to be recalled. 
Calude and Longo showed that spurious correlations are a consequence purely of the size, not the nature, of a data set \cite{Calude}. 
Such multitudinous spurious correlations also contribute to the low value of these exponents; they are intrinsic and overwhelmingly predominate over true correlations as the size of these data sets increases. It is clear that the mainstream literature on AI has completely overlooked this finding, although it is central to the problems we encounter in LLMs and indicates that they cannot be eradicated by mere brute force (see Appendix III).

\subsection{Back to the exponents}

The above matters had already been anticipated by the present
authors back in 2019 \cite{SC19}, although in much less quantitative terms and
without any specific reference to actual data, since such data 
were not available at the time of that writing.
Since these data are now available, it is worth inspecting a few elementary 
but far-reaching consequences of the resilience of uncertainty {\it specifically 
associated with the most optimistic scaling exponent of $\alpha=0.1$}.

For this purpose, let us recall that what a Monte Carlo exponent of $1/2$ means 
in actual practice is that, in order to cut down the error in a calculation by a factor $10$, one 
needs to allocate $10^2=100$ more compute resources.
This is usually regarded as a tough scaling law to live with in scientific
applications, but it frequently offers the only way to go with 
ultra-high-dimensional systems. For the case of $\alpha=0.1$, reducing the error 
by a factor $10$ requires the allocation of $10^{1/0.1}=10^{10}$, i.e. ten 
billion, more compute resources.
And, if we focus on the power consumption itself, the exponent is $0.05$, meaning 
an astronomical $10^{20}$ more computing power for just one order better in accuracy. 

Despite the hype which seeks to persuade us that present-day LLMs already match
and even outdo humans, it is acknowledged by the leading AI tech companies themselves that 
while LLMs produce ``answers" very quickly, they make many mistakes, 
never mind their propensity to hallucinate \cite{Wachter}.
%{\bf Sandra Wachter, Brent Mittelstadt  and Chris Russell, ``Do large language models have 
%a legal duty to tell the truth?" Royal Society Open Science (2024) https://doi.org/10.1098/rsos.240197}, 
That is why these AI companies have been trying to build increasingly large models and have 
now begun to openly recognise that the paradigm is nowhere near 
the nirvana of artificial general  intelligence which they not long ago proclaimed to be \cite{Aschen}.
%{\bf Leopold Aschenbrenner's ``Situational Awareness: 
%The Decade Ahead" https://situational-awareness.ai (2024)}.
Instead, a new form of AI is heaving into view - so-called {\it agentic AI}.

The main point here is that the scalability characteristics of these algorithms 
carries within them the seeds of their future failure.  
Even assuming that the original optimism behind LLMs were to be maintained, the scope 
for improvement is absolutely untenable on account of the accuracy required 
for most scientific applications \cite{Mc}, let alone the power demands of the approach. 

% -------------------
\begin{figure}
\centering
\includegraphics[scale=0.3]{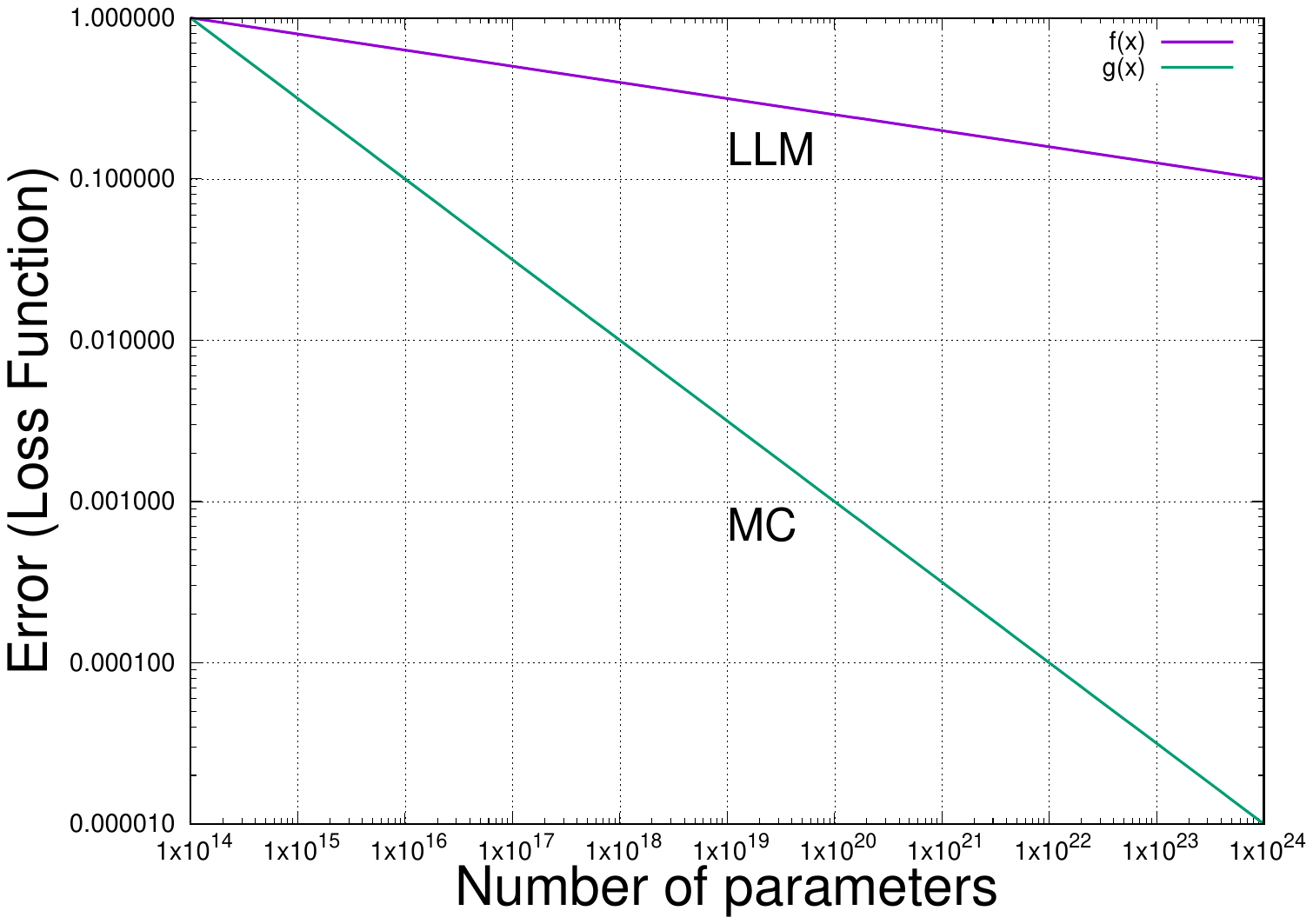}
\caption{LLM versus Monte Carlo scaling. Starting from $N=10^{14}$,
it takes another $10$ decades, $N=10^{24}$ to reduce the LLM error
by a factor 10. The MC scaling achieves the same factor 10 with just 
two decades, $N=10^{16}$. In fact, the true gap is much wider than this, since
a more realistic value for MC is $G_c \sim 10^9$.} 
\end{figure}
% -------------------

%{\bf On my print-out I cannot see any trace for the 
%MC behaviour in this figure -- maybe the choice of colour is too 
%faint to be visible? 
%SS: it shows fine by me, i can thicken it though
%}

Again, we wish to emphasise that this behaviour is fully in line with the expectation that, due to their non-Gaussian stochastic nature, LLMs should indeed retain an extreme form of uncertainty in their predictions which decays only very slowly regardless of the amount of data upon which they are trained.
This is basically the main message broadcast by exponent below 
$0.1$ (see also below). Although our article is based on the 2020 paper by Kaplan et al., subsequent
work by DeepMind with larger datasets \cite{Hoffmann} conveys 
essentially the same message; see also the independent assessment in \cite{Wolfe}.
In this paper, Wolfe provides a very informative assessment of scaling from
a variety of angles; to be sure, the central point is definitely not missed, as can 
be noted in comments such as ``The idea of getting exponential scaling laws from logarithmic increase
of compute is a myth" and ``Scaling will eventually lead to diminishing returns".
Nevertheless, these statements do not do justice to the magnitude of the problem exposed by the actual numbers, which is the main purpose of our current paper.  

\section{The wall}

Despite the major technological hype, computer scientists working on AI are well aware
of the significant error rate incurred by even the most advanced chatbots,
a rate which is definitely incompatible with the large majority 
of scientific applications. 
This is why there is currently a heated debate about the future of scalability and ways 
to overcome the wall that, even now, stands in the way of a golden future for LLMs. 

Formally, walls are signalled by a sign change of the exponent, reflecting
{\it negative} return on investment: by increasing the resources
the accuracy does not get any better but actually much worse.
Paradoxical as it may seem, this is a plausible outcome, as it
reflects the onset of adverse effects that remain silent and out of view 
below the wall threshold.
Again, the simple example of the spatial derivative offers clearcut insight.
As one learns at high school, the continuous derivative is the limit of the
discrete version as the displacement $h$ is sent to zero. 
If our computers could afford infinite precision, this statement would be equally 
good in practice as it is in continuum mathematics.
But no computer can afford infinite precision, in fact, the standard double-precision
IEEE representation of floating numbers offers an accuracy around the 16th digit,
meaning that numbers below $10^{-16}$ are basically treated as pure noise.
This means that upon sending the displacement $h$ below machine precision, the 
discrete derivatives start to diverge from the continuum value 
as roundoff errors then dominate the discretization errors.  
The two curves shown in Fig 2 provide an excellent example of the roundoff 
wall in practice for single and double precision floating point numbers.

% -------------------
\begin{figure}
\centering
\includegraphics[scale=0.3]{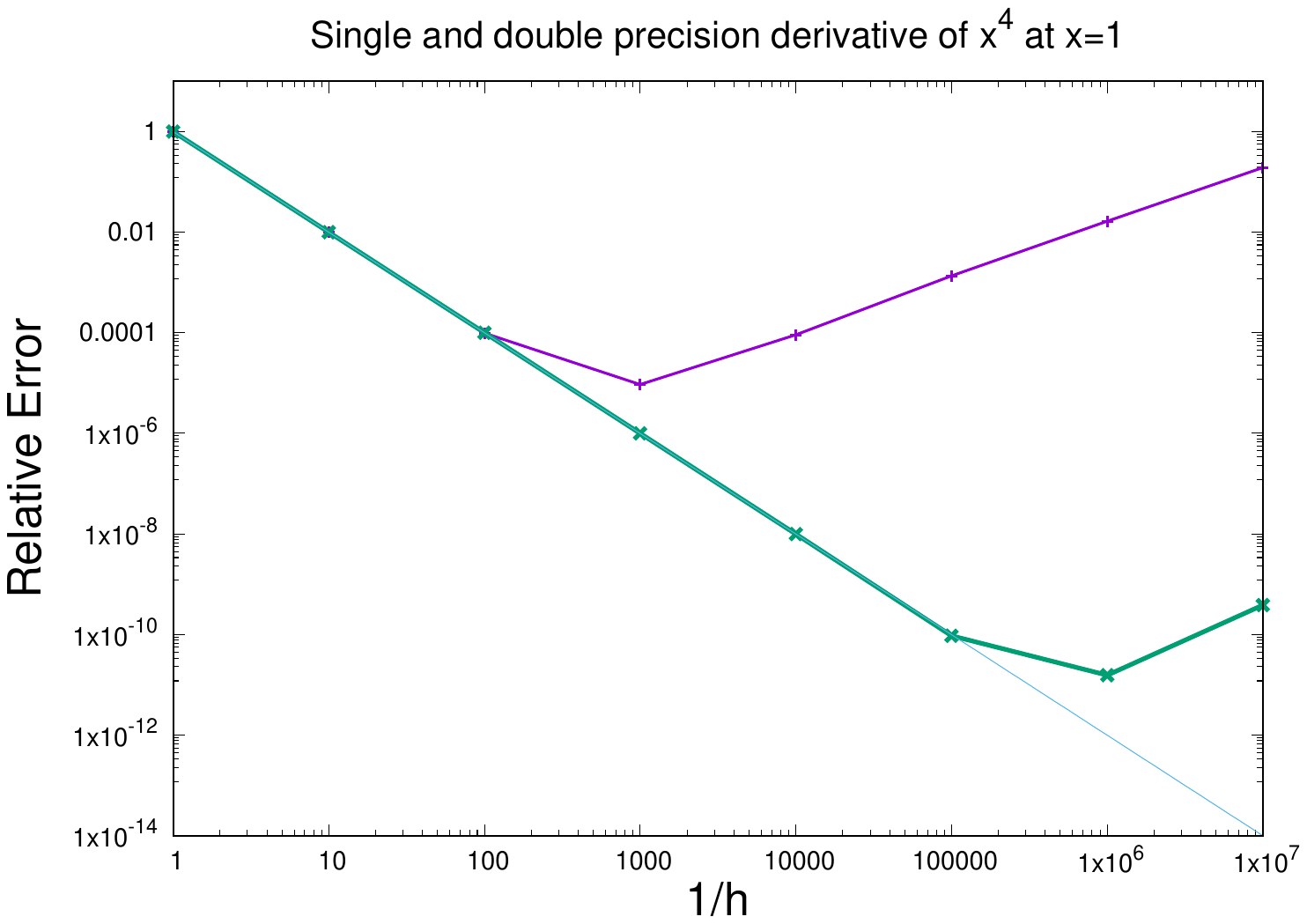}
\caption{The absolute value of the relative error of the 
second-error accurate discrete derivative of $f(x)=x^4$ 
at $x=1$ at decreasing values of the increment $h$. 
The upper curve, in single precision, adheres to the
analytical scaling $h^2$ (thin straight line underneath)
until it hits the roundoff wall at $h \sim 10^{-3}$
consistent with the 7 digit accuracy of $h^2$. The same derivative in 
double-precision pushes the wall out to $h \sim 10^{-6}$.   
}
\end{figure}
% -------------------

Of course, LLMs and their upgrades, LRMs and agentic AIs, are immeasurably more complex than a simple derivative. 
Consequently many potential walls present obstacles in their way as their size increases.  
Yet, our simple example is representative, as it reminds us of another {\it caveat}, namely
that digital systems are exposed to roundoff errors and increasingly so 
as their complexity and the number of operations required to complete their tasks increase. 
This issue has been recently raised with regard to the simulation of 
chaotic systems \cite{BCW,PVCshark} and certainly deserves close
attention for LLMs as well.   
That said, we wish to return to the main point of this article, namely the fact 
that LLM scaling exponents, albeit positive (wall-free, for now), 
are simply too small to be sustainable beyond the present.
In the light of this observation, the debate as to whether LLM scaling 
is going to hit a wall appears to us to be a comparatively marginal one.

%This is what we mean by ``transhuman scaling": one that cannot 
%be sustained by humankind as a whole, but only (perhaps) by a hipertiny 
%elite willing to exhaust the energy resources of the planet 
%in the name of economic efficiency.  

\section{Accuracy and reliability}

It is generally agreed that the accuracy of machine learning applications
is highly dependent on the homogeneity of the training data set and
can deteriorate substantially once exposed to unseen data drawn 
from a dataset other than the one used for 
training and validation \cite{Guillo}. 

Before commenting on this issue, let us first remark that
issues of accuracy often arise even within homogeneous training scenarios.
Although the details may change depending on the specific context, a few 
general traits which hamper accuracy can be identified.

A primary one occurs within {\it multiscale physics}, as a direct consequence
of strong non-linearity, strong inhomogeneity or both of these.
Whenever interactions take place simultaneously on different scales
of motion, learning the dynamics at one scale does not necessarily
imply transferability to another scale. This phenomenon has been noted in the application of 
ML-based diffusion models to the problem of Lagrangian turbulence
\cite{Bifer}, characterized by heavy-tailed distributions of the
particle trajectories, resulting in a scale-dependent degree
of non-Gaussianity (for a quantitative definition of non-Gaussianity
see Appendix I). In other words, the problem is not
non-Gaussianity {\it per se}, but a {\it changing degree of non-Gaussianity
with scale} \cite {biferale2025private}.
%{\bf I can ask a beautiful pic to Biferale}
 
In a similar vein, if the dynamics is chaotic on one scale there are 
lots of discontinuities in the dependence of the physical 
system on the parameters of the real system, which are 
overlooked by the smoothing caused by most ML training algorithms,
especially whenever the latter are developed in the
formulation of low-dimensional surrogates.
Such discontinuities, or more generally, sharp features, are inherently
exposed to rapid amplification due to the chaotic nature of 
the dynamics \cite{PCSW}. This is not to say that such exposure
will {\it invariably} result in untamed error amplification but the 
evidence is that at a minimum particular forms of trained networks are 
then required, in particular so-called {\it physics-informed} ones.
 
Let us now return to the heterogeneous scenario. 
Although seldom explored in quantitative terms, this 
lack of generalizability remains a major issue for machine
learning, because generalizability is {\it de facto} just a glorified
word for the basic essence of learning itself. 
Validation, which in this context means predicting unseen data from the 
same ``held-out" dataset used for training, is of distinctly limited value as 
it is likely to be more of a memorization process than a genuine learning one.
 
In this respect, even though many deep learning
applications, particularly LLMs, have been hailed for their 
unanticipated capability to ``predict the unseen" and more generally to produce 
plausible responses like those of humans, this cannot be used 
as an excuse for glossing over the issue of the 
reliability of such predictions. 
All the tech companies' AI tools may indeed keep surprising us, but this does not change 
the fact that they still make many mistakes (and indeed openly state 
as much in the small print they provide to users), far too many for the 
standards of accuracy required for most scientific and many other professional 
domains such as the legal and educational fields \cite{Wachter}. 
%{\bf Cite again:  Sandra Wachter, Brent Mittelstadt  and Chris Russell, 
%"Do large language models have a legal duty to tell the truth?" 
%Royal Society Open Science (2024) https://doi.org/10.1098/rsos.240197}.

AlphaFold earned a Nobel prize in Chemistry for its prowess in protein structure prediction, 
but the fact remains that it is still exposed to substantial 
failure rates when faced with unseen structures from different
scientific sources than those it was trained on. 
In particular, AF was trained with both structural information from existing 
PDB structures and multiple sequence alignments.  
This generates the predicted models which are refined with the Amber molecular 
modelling software in later stages of the algorithm.  
There is no surprise that this explains why the predicted models tend to be good 
when used to predict structures with similar high sequence homology to those in the training data. 
In short, AlphaFold predictions are valuable hypotheses and accelerate but do not in general
replace experimental structure determination.
(The limitations with respect to experimental structures -- i.e. validation -- can be 
found in \cite{terwilliger2024alphafold}).

\section{Beyond LLMs: LRMs and Agentic AI}

The limitations of LLMs in terms of their reliability and power consumption are 
well known, but rarely articulated in any technical detail, especially for general 
public consumption. Indeed, their discovery and development have followed a highly empirical pathway, which in many respects resembles trial-and-error. As noted in \cite{Wolfe}, the scaling
properties discussed this far refer primarily to the training era. 
Given the lack of general reliability of their outputs, the high-tech industry is moving 
to enhance their credibility not solely by improving the amounts of data that 
they are trained on, but changing the kind of AI being used. We have already mentioned the advent of LRMs whose empirical basis makes it even harder to quantify performance effectively.
Indeed, {\it Agentic AI} has also now been brought into play. 
Rather than relying on the effectiveness of one single AI system to handle all 
forms of query thrown at it, the idea is now to seek to improve the scalability 
and the quality of output information by adapting their architecture and deployment 
in order to handle larger volumes of data, more complex tasks, and to access larger 
user databases through the participation of a proliferation of 
individual AI systems, each of which one can think of as an agent \cite{Albrecht}. However, agents are about getting LLMs to do more than chat; supposedly, in this manner they should create economic value.They provide a useful narrative from a business or product perspective but lack any means of serious scientific assessment, being a catch-all term that encompasses reasoning, multimodality, continual learning, swarm intelligence and whatever comes next. 

This overall aim is of course very sensible: such AI agents should be capable
of developing human-like ``reasoning attitudes", such as pondering before 
answering which is one such trait they exhibit (even if not all humans do this).  
The hope is that  such an approach is going to dramatically cut down their error 
rate and enhance their reliability, even if it involves waiting longer to get an answer.  
This is the basic idea behind what is known as the Chain of Thought (CoT), a strategy 
mirroring human reasoning, aimed at facilitating systematic 
problem-solving through a coherent series of logical deductions \cite{YuWang}. 
The simple question is: will all of the above lead to sustainable, 
scalable strategies? 
Insofar as agentic AI depends on components which display the features discussed here that are characteristic of LLMs, this seems highly unlikely.    
Perhaps a more productive way forward would be to let LLMs do what generative models are meant to do: hallucinate. Reasoning models and multi-turn tool use can be seen as productive steps in this direction: the LLM proposes the next step, while other components of the system handle evaluation and reward optimisation. In this setup, hallucination is not suppressed but channelled, turning generative looseness into exploratory value. AlphaEvolve follows a similar strategy, using an LLM to dream up code variants, substituting reinforcement learning with an evolutionary algorithm to guide selection and refinement \cite{xia2025private}.

%(for an informed critical assessment see \cite{Guttmann}) {\it It would be preferable 
%if we could cite the original works where we can rather than a LinkedIn posting.} 

\section{Conclusions: an avoidable path to Degenerative AI}

The considerations collected in this work configure a potential pathway to a
doom-laden scenario which has been named Degenerative AI \cite{Hammond}, meaning the
catastrophic accumulation of self-feeding errors and inaccuracies, which appear particularly
likely in the case of LLMs trained on synthetic data.

Although we do not have the slightest desire to ride any ``catastrophic
wave”, it is reasonable to point out not only the plausibility but even the inevitability of the DAI scenario based
on the theoretical considerations articulated here. The DAI causal chain runs
as follows: low scaling exponents are the smoking gun of non-Gaussian fluctuations,
which support anomalous resilience of uncertainty, thereby opening
the door to the untamed pileup of errors caused by an inability to accurately
represent the tails in the distributions, ultimately ending up in an information
catastrophe \cite{Mitchell}.
Symbolically, the DAI chain reads as follows:
\begin{equation}
\label{DAI}
SSE : NGF \to RoU \to IC
\end{equation}
where $SSE$ stands for ``Small Scaling Exponents", $NGF$ for ``Non Gaussian Fluctuations",
$RoU$ for ``Resilience of Uncertainty" and $IC$ for ``Information catastrophes". 
The fact that such a DAI pathway is an intrinsic element of the existing LLM landscape by no means implies that it must be taken. 
Data are often implicitly assumed to be the same as information but this is obviously not true, as has been noted by several authors besides Calude and Longo \cite{Chibbaro2017}. Indeed, there are several mechanisms by which adding more data can decrease the amount of information, as is the case whenever data conflicts arise or false data are deliberately injected to produce fake news 
(a process called ``poisoning").
The fact that the scaling exponents, albeit very small, are still positive indicates that the degenerative
regime whereby more data imply less information has not yet been encountered. 
However, as repeatedly noted in this work, it implies a regime of strongly diminishing returns.
With a putative scaling relation of the form 
$$
I = (D/D_c)^{0.1}
$$
where $I$ is information and $D$ is data size, if data were to double every year, information would double only every ten years.
Even assuming the so-called Huang's law, namely doubled GPU power every year, this factor two would still take ten years. 
The Calude-Longo finding bears strongly on this regime of highly diminishing returns from scraping more data together because then the spurious correlations overwhelm the true ones. In a way, one could say that the low scaling exponents are the result
of a near-tie between the exponential growth of resources and the exponential decay of the true correlations with data size.
What it means is plainly that sacrificing understanding and insight on the altar of brute force and 
unsustainable scaling boosts the chances that it will indeed persist no matter how big the data. 
By contrast, the scientific method provides precisely the means to sieve off the true correlations from the vastly greater sea of spurious ones through the construction of ``world models”\cite{vafa2025has}. Simply ignoring it on the supposition that brute force can solve the problem is doomed to failure.

\section{Acknowledgements}
We are grateful to W. Edeling, A. Laio, Y. Yang, D. Spergel, F. Xia, N. Chia, N. Liu and W. Keyrouz for many enlightening discussions.
SS thanks SISSA for financial support under the ``Collaborations of Excellence" initiative as well as the Simons Foundation for supporting several enriching visits. The same author acknowledges financial support from the EIC ``iNSIGHT" Project.
Likewise PVC thanks EPSRC for funding under grant number EP/W007711/1.  

\newpage 
\section*{Appendix I: Resilience of Uncertainty in LLMs}

In this paper we have repeatedly mentioned resilience of uncertainty (RoF)
without specifying in detail what exactly we mean by this expression.
Here we discuss the concept is some detail.
The uncertainty associated to a given stochastic 
variable $x$, is encoded within the central moments of its 
probability distribution
\begin{equation}
M_k = <(x-\mu)^k> = \int p(x) (x-\mu)^k dx
\end{equation}  
where $\mu = <x>$ is the mean.
The most popular one is the variance $\sigma^2 \equiv M_2$
which measures the Euclidean distance between the 
dataset of realizations $\lbrace x_i \rbrace, i=1,\cdots ,n$
where $n$ is the number of samples.
The standard deviation $\sigma=M_2^{1/2}$ provides a direct
measure of the uncertainty associated with a measurement 
of the stochastic variable $x$, which is estimated as
$\mu \mp \sigma$.
This estimate is particularly suited to Gaussian statistics
controlled by the Gaussian distribution 
$p_G(x) = (2 \pi)^{-1/2} e^{-y^2/2}$, where $y=(x-\mu)/\sigma$
is the de-trended observable.
However, higher order metrics could be used as well, reflecting
the idea of giving more weights to outlier events, with $x \gg \sigma$.
For instance, for the Gaussian distribution,
the case $k=4$ delivers $M_4=3 M_2^2$, which is higher than $M_2^2$
precisely because of the enhanced weight of events with 
$x \sim \sqrt{4} \sigma$ (a moment of order $k$ gives most of
the weight to event with $x/ \sigma \sim \sqrt{k}$.
The standard deviation of order 4 can then be defined as
$$
\sigma_4^4 = 3 \sigma^4 ,
$$    
delivering $\sigma_4 = 3^{1/4} \sigma \sim 1.73 \sigma$, almost
twice the usual one.
By the same token, the standard deviation at order $k$  ($k$ even)
is given by 
\begin{equation}
\sigma_k^k = m_k \sigma^k ,
\end{equation}
where we have defined $m_k = M_k/(M_2)^{k/2}$.
That is:
\begin{equation}
\label{SIGMAK}
\sigma_k = (m_k)^{1/k} \sigma \equiv a_k \sigma .
\end{equation}
where $a_k$ are the amplification coefficients of uncertainty
at various orders.  

For a Gaussian distribution, this sequence can be 
computed analytically, delivering $m_k = (k-1){!!}$,
that is, $m_4 = 3$, $m_6 = 5 \times 3=15$, 
$m_6=7 \times m_4 = 105$, $m_8 = 9 \times m_6=945$
and so on. From eq.(\ref{SIGMAK}) it is rapidly seen that
the amplification of uncertainity due to the outliers
grows very slowly with the order of the moment. 
For the first ten moments, we have
$a_4 = 1.73$, $a_6 \sim 1.57$, $a_8 \sim 1.79$
and $a_{10} \sim 1.98$.
This slow growth is precisely the opposite of what we call 
Resilience of Uncertainty. Small amplification 
factors reflect strong suppression of outliers and therefore low resilience to uncertainty, hence the sequence of amplification
factors provides a well-defined and quantitative measure of RoU. 
Once again, this is just another way of stating that outliers are
heavily suppressed in the Gaussian world.

But what about non-Gaussian distributions, such as stretched exponentials
or power-law distributions, which are a commonplace in 
the statistical dynamics of complex systems \cite{Balescu}? To state the case, let us consider the extreme case of the (detrended) Lorenz-Cauchy distribution 
$$
p_L(x) = \frac{1}{\pi} \frac{1}{1+x^2}.
$$
If the stochastic variable is allowed to take infinite values
on both sides, the variance is infinite, which undermines
the RoU sequence from the very beginning. Since no real data can admit infinite values, let us 
consider the finite-size case $|x|<L$, with $L \gg 1$
so as to allow {\it extreme} events with $x \gg 1 $
(particle trajectories in turbulent flows can reach up to 50
standard deviations \cite{Bifer}).

On the assumption of $L \gg 1$, the variance $M_2$
is readily estimated to scale like $L$, whereas $M_4 \sim L^3$.
As a result, we have $a_4 \sim L^3 /L^2 = O(L)$.
Likewise $M_6 \sim L^5$, so that $a_6 \sim L^2$ and so on.
Hence, the RoU coefficients are still finite, but they
diverge with increasing powers of $L$, the maximum intensity of the process.
With a realistic values $L=100$, the standard deviation at level
$4$ is 100 times larger than the usual one.
This clearly shows what Resilience of Uncertainty means in actual
practice. The implication concerning the required size of the sample is immediate: since
the standard deviations of higher orders have been based on the
standard one with $k=2$, we can still apply the Gaussian
$1/\sqrt{n}$ estimate, except that $\sigma$ is now one hundred times larger.
Hence, in order to achieve the same accuracy one would obtain for
a Gaussian process, one needs $100^2$ more samples. This is the effect of rare-events with high intensity, which are no longer heavily suppressed as in the Gaussian world.

\section*{Appendix II: Non-Gaussian Distributions}

Let us consider the transformer equation (\ref{eq:dzdt}) for the case of two
neurons $N=2$ and just an input and an output layer. 
The  transformation reads explicitly as follows:
\begin{eqnarray}
y_1 = f(W_{11}x_1+W_{12}x_2)\\
y_2 = f(W_{21}x_1+W_{22}x_2)
\end{eqnarray}
The Jacobian of the transformation is given by:

\begin{eqnarray}
J_{11} \equiv \frac{\partial y_1}{\partial x_1} = f'(z_1) W_{11}\\
J_{12} \equiv \frac{\partial y_1}{\partial x_2} = f'(z_1) W_{12}\\
J_{21} \equiv \frac{\partial y_2}{\partial x_1} = f'(z_2) W_{21}\\
J_{22} \equiv \frac{\partial y_2}{\partial x_2} = f'(z_2) W_{22}
\end{eqnarray}
where we have set $z_i = W_{ij}x_j,\;i,j=1,N$ and
$f'$ denotes the derivative of the activation function 
with respect to its argument $z_i$.
As a result 
$$
|det[J]| = |det[W]| |f'(z_1)f'(z_2)|
$$
Assuming an independent joint Gaussian distribution for the input data with mean
$<x_i>=m_i, \;i=1,2$ and variance $\sigma^2$, 
$$
p_X(x_1,x_2) = \frac{1}{\sqrt {2 \pi} \sigma^2} 
e^{-\frac{(x_1-m_1)^2+(x_2-m_2)^2}{2 \sigma^2}}
$$
the output distribution is given by
$$
p_Y(y_1,y_2)) = \frac{1}{\sqrt{2 \pi} \sigma^2} 
e^{-\lbrace \frac{(x_1-m_1)^2+(x_2-m_2)^2}{2 \sigma^2} + log (|det[J]|) \rbrace}
$$
where $x$ as a function of $y$ is obtained by inverting the
transformer equation, formally:
$$  
x_i = W^{-1}_{ij}f^{-1}(y_i),i,j=1,{\cdots}, N.
$$
This shows that, unless $f(z)$ is linear, the 
the Jacobian is not a constant, hence the output 
distribution is non-Gaussian.
Of course, non-Gaussianity does not necessarily imply 
a fat-tailed distribution, the latter feature depending on 
the specifics of the neural nets. 
Yet, some general criteria are easily spotted. For instance by using
$f(z)= tanh(z)$ it is clear that small weights, $W \ll 1$ tend to produce
small $z=Wx$, hence $y \sim f(0) =0$, giving rise to a delta like distribution
$p_Y(y) = \delta(y)$. On the contrary, large weights $W \gg 1$, tend to produce large 
$Wx$, hence $y=tanh(z) \sim \pm 1$, leading to a bimodal distribution
$p_Y(y) = \frac{1}{2} (\delta(y+1)+\delta(y-1))$.
Clearly, the transition between these two extremes is characterised by flat 
distributions, see Fig.3.
% -------------------
\begin{figure}
\centering
\includegraphics[scale=0.3]{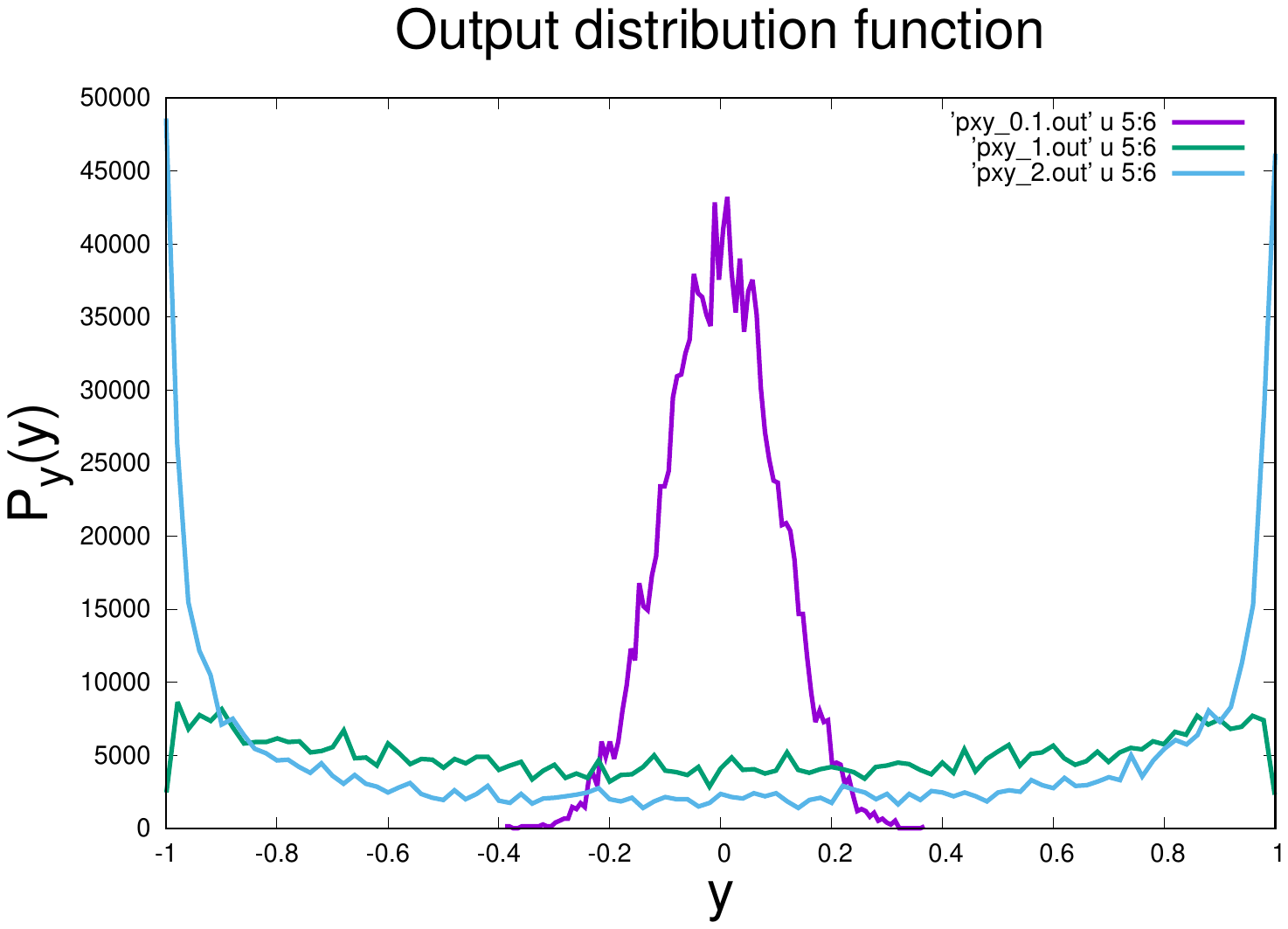}
\caption{The output distribution function $p_Y(y)$ generated 
by a normal distribution $p_x(x)$ using $y=tanh(Wx)$ wit:x
h $W=0.1,1,2$, respectively
The transition from a Gaussian shape for $W=0.1$ to a double humped shape at $W=2$
is clearly visible.
The distributions are generated with $10^4$ samples.
}
\end{figure}
% -------------------

\section*{Appendix III: Calude-Longo deluge of spurious correlations}

The main power of insight is the resulting ability to describe complex phenomena 
with less information than needed otherwise.
Symmetries are a major example in point: knowing that a function is even, $f(-x)=f(x)$,
halves the information needed to fully describe it.
Calude and Longo start by defining a spurious correlation as one that can be generated 
by a random process, regardless of its apparent regularity.\footnote{
As a simple example, consider the 16-bit string $0010100111011001$, which shows 
some regularity in the form of two identical 4-bit substrings $1001$, and three identical
3-bit substrings $001$, despite having been generated by a fully random procedure
(draw a uniform random number $0 \le r \le 1$ and set the bit to $0$ if $r \le 1/2$ and $1$ otherwise).
This is not surprising, since the TC/FC ratio in this case is approximately $2^{-0.13 \times 16}  \sim 0.23$.
} 
The authors then show that such spurious correlations are exponentially more numerous than the true ones.
To prove this result, they consider a $n$-digit long binary string $s$ and observe that in order to 
be compressible by $\kappa n$ bits, from $n$ to $n-\kappa n$, the complexity of the string, defined as the
amount of information required to exhaustively describe all string configurations, must obey the inequality
$$
C(s) \le (1-\kappa)n 
$$ 
The authors prove that the number of such strings is $2^{(1-\kappa) n}$
over a total of $2^n$, so that the ratio $2^{-\kappa n}$ decays exponentially with
the string length. Note that this result is fully general, i.e. it depends only on the
string length (size of the dataset) and the compressibility index $\kappa$.
Independent studies show that the latter can hardly exceed the value $\kappa \sim 0.13$,
meaning that the ratio of true to spurious correlations (TC/FC, where TC stands for true and F for false, i.e. spurious, correlations) for a comparatively short string of
1024 bits is approximately $2^{-130} \sim 10^{-39}$!
This is the Calude-Longo ``deluge" of spurious correlations, a deluge largely ignored by the
AI mainstream, for no evident good reason.

\newpage 
\bibliographystyle{plain} 
\bibliography{bibML2}
\end{document}